\documentclass[preprints,article,accept,pdftex,moreauthors]{Definitions/mdpi}

\firstpage{1} 
\pubvolume{1}
\issuenum{1}
\articlenumber{0}
\pubyear{2022}
\copyrightyear{2022}
\datereceived{} 
\dateaccepted{} 
\datepublished{} 
\hreflink{https://doi.org/} 
\pdfoutput=1

\Title{UNav: An Infrastructure-Independent Vision-Based Navigation System for People with Blindness and Low vision}

\TitleCitation{UNav: An Infrastructure-Independent Vision-Based Navigation System for People with Blindness and Low vision}

\Author{Anbang Yang $^{1}$, Mahya Beheshti, MD $^{1,2}$, Todd E Hudson $^{2}$, Rajesh Vedanthan $^{3}$, Wachara Riewpaiboon $^{4}$, Pattanasak Mongkolwat $^{5}$, Chen Feng $^{1}$* and John-Ross Rizzo, MD, MSCI $^{1,2}$*}

\AuthorNames{Anbang Yang, Mahya Beheshti, Todd E Hudson, Rajesh Vedanthan, Wachara Riewpaiboon, Pattanasak Mongkolwat, Chen Feng and John-Ross Rizzo}

\AuthorCitation{Yang, A.; Beheshti, M.; Hudson, T.; Vedanthan, R.; Riewpaiboon, W.; Mongkolwat, P.; Feng, C.; Rizzo, JR.}

\address{%
$^{1}$ \quad NYU Tandon School of Engineering, NY, USA; cfeng@nyu.edu\\
$^{2}$ \quad Department of Rehabilitation Medicine, NYU Grossman School of Medicine, NY, USA JohnRoss.Rizzo@nyulangone.org\\
$^{3}$ \quad Department of Population Health, NYU Grossman School of Medicine, NY, USA\\
$^{4}$ \quad Ratchasuda College Mahidol University, wachara.rie@mahidol.ac.th\\
$^{5}$ \quad Faculty of Information and Communication Technology, Mahidol University, Salaya, Nakhon Pathom 73170, Thailand;
}

\corres{Correspondence: cfeng@nyu.edu, johnrossrizzo@gmail.com}

\abstract{Vision-based localization approaches now underpin newly emerging navigation pipelines for myriad use cases from robotics to assistive technologies. Compared to sensor-based solutions, vision-based localization does not require pre-installed sensor infrastructure, which is costly, time-consuming, and/or often infeasible at scale. Herein, we propose a novel vision-based localization pipeline for a specific use case: navigation support for end-users with blindness and low vision. Given a query image taken by an end-user on a mobile application, the pipeline leverages a visual place recognition (VPR) algorithm to find similar images in a reference image database of the target space. The geolocations of these similar images are utilized in a downstream task that employs a weighted-average method to estimate the end-user's location. Another downstream task utilizes the perspective-n-point (PnP) algorithm to estimate the end-user's direction by exploiting the 2D-3D point correspondences between the query image and the 3D environment, as extracted from matched images in the database. Additionally, this system implements Dijkstra’s algorithm to calculate a shortest path based on a navigable map that includes trip origin and destination. The topometric map used for localization and navigation is built using a customized graphical user interface that projects a 3D reconstructed sparse map, built from a sequence of images, to the corresponding a priori 2D floor plan. Sequential images used for map construction can be collected in a pre-mapping step or scavenged through public databases/citizen science. The end-to-end system can be installed on any internet-accessible device with a camera that hosts a custom mobile application. For evaluation purposes, mapping and localization were tested in a complex hospital environment. The evaluation results demonstrate that our system can achieve localization with an average error of less than $1$ meter without knowledge of the camera’s intrinsic parameters, such as focal length.}


\begin{document}

\begingroup
\let\clearpage\relax
\section{Introduction}

According to the International Agency for the Prevention of Blindness, 295 million people are presently living with moderate-to-severe visual impairment and 43 million are living with blindness, a number projected to reach 61 million by 2050 \cite{kruk2020lancet}. Vision loss disproportionately deprives multi-sensory perception when compared to other sensory impairments and has been shown to significantly decrease mobility performance or the ability to travel safely, comfortably, gracefully, and independently through the environment \cite{hakobyan2013mobile} Consequently, people with blindness and low vision (BLV) have difficulty travelling efficiently and finding destinations of interest \cite{kruk2020lancet}.

Since the 1960s, numerous assistive technologies have emerged \cite{kandalan2019comprehensive} to tackle travel difficulties. These technologies target context-awareness in the form of vision replacement, vision enhancement, and vision substitution \cite{dakopoulos2009wearable}. The focus of this paper is vision substitution, for which three subcategories of devices exist: Position Locator Devices (PLDs), Electronic Travel Aid (ETAs), and Electronic Orientation Aid (EOAs). Most of the commercial offerings in these categories have yet to gain significant market traction due to low accuracy, cost, and feasibility / implementation barriers, such as the need for physical sensor infrastructure. 

This paper proposes a novel sensor-infrastructure-independent system for assistive navigation; the approach is cost-efficient and highly accurate with an average error of less than $1$ meter. Our system is based on topometric maps computed by simultaneous localization and mapping (SLAM) and structure from motion (SfM) algorithms. One distinct advantage of our system is a map-evolution feedback loop, in which query images from the target space are re-directed into a reference image database, accounting for dynamic changes in the target space and improving the density of the map data. Our system uses visual place recognition (VPR), weighted averaging and perspective-n-point (PnP) algorithms for localization. More specifically, we adopt NetVLAD \cite{arandjelovic2016netvlad} for global descriptors and Superpoint \cite{detone2018superpoint} for local descriptors to aid in the localization process. Once the localization is rendered, a suggested destination can be entered into a navigation pipeline and directions are generated. Navigation instructions are computed using the Dijkstra algorithm and based on connecting the end user's current location with a destination of interest. The system runs on a cloud server, which receives data as well as input commands and sends navigation instructions to the end-user’s preferred mobile device over the internet. In cases of signal loss, our solution supports offline computation locally on the end-user device. This paper will discuss two types of end-user devices that we developed. One is an Android app based on Java language, and another is a backpack system with an Nvidia Jetson AGX Xavier and a bone-conduction headset.

The remainder of this paper is arranged as follows: a related work section about sensor-based and vision-based navigation systems, a methods section that describes our system architecture and two end-user devices, an evaluation / results section, and, lastly, a discussion and conclusion section.

\section{Related Work}

Over the past three decades, many assistive technologies (AT) have been developed to help the BLV navigate independently and safely in unfamiliar environments \cite{manjari2020survey}. These AT focused on navigation can be broadly divided into two groups: \textit{sensor-based} and \textit{vision-based}.

\textbf{Sensor-based devices} which are dependent on pre-installed sensor input are potential solutions for navigation pipelines. However, all sensor-based technologies when translated at-scale, ensuring entire spaces are accessible, suffer from logistical issues, most importantly unrealistic economics. Devices that use Wi-Fi \cite{yang2021decimeter}, Bluetooth \cite{al2019fuzzy} or ApriTag \cite{feng2012augmented} require the pre-installation of beacons/modules and tedious calibration routines, driving up cost, maintenance, and inaccuracy. To tackle these issues, vision-based devices have been developed. Most use a smart mobile device equipped with a camera, as a cost-efficient input sensor. Vision-based localization has two subcategories: \textit{retrieval-based} localization and \textit{pose-based} localization. 

\textbf{Retrieval-based} localization, also known as image-based localization, uses a visual place recognition (VPR) algorithm to retrieve a set of reference images from a database that are visually similar to a query image taken by an end-user, whose location can be estimated by extracting and averaging the geolocation of the retrieved reference images. The geolocations can be either obtained from GPS or a 3D-reconstructed model. The VPR algorithm has two steps: \textit{feature aggregation} and \textit{similarity search}. 

\textbf{\textit{Feature aggregation}} aims to represent an entire image as a low-dimensional vector assembled from the image's feature points in order to accelerate the searches when matching database images to a query image. BoVWs \cite{csurka2004visual,sivic2003video}, VLAD \cite{jegou2010aggregating}, and DenseVLAD \cite{torii201524} are three traditional handcrafted feature aggregation algorithms that determine feature points by exploiting relations between each pixel of the image and its adjacent pixels. In 2016, NetVLAD proposed to use VLAD \cite{jegou2010aggregating} in an end-to-end trainable deep neural network, which instead extracts feature points implicitly by a trained network. A series of evaluations has shown that NetVLAD outperforms the traditional handcrafted methods by a significant margin \cite{arandjelovic2016netvlad}.

\textbf{\textit{Similarity search}} aims to find the similar reference images by isolating those whose low-dimensional vectors have minimal distances (e.g., Euclidean) to the query image’s vector through an exhaustive search. However, this search may be computationally expensive when the reference image database becomes large. To tackle this problem, the nearest-neighbor search method was proposed to reorganize the data's store structure to speed up searching, as employed in the K-D tree \cite{bentley1975multidimensional}, a hash table \cite{gennaro2001similarity}, or quantization frameworks \cite{philbin2007object,nister2006scalable}, trading accuracy for rapidity.

\textbf{Pose-based} localization, unlike retrieval-based localization, calculates the more accurate 6 DoF pose of the query image relative to the 3D space. There are three approaches in this class. 

\textbf{\textit{The first approach directly regresses the pose from a single image using a deep neural network \cite{kendall2015posenet,kendall2017geometric,brahmbhatt2018geometry,wang2021deep}.}} The network represents implicitly a 3D reconstruction of the target space, to retrieve the image’s pose. Evaluations have shown that, despite being efficient, the localization of this approach is often inaccurate \cite{Kendall_2017_CVPR}.

\textbf{\textit{The second approach retrieves the query image pose by leveraging coarse prior information.}} This approach focuses primarily on refining the estimated coarse camera pose using pre-known geo-information obtained by GPS, Wifi, Bluetooth, a reconstructed 3D map, etc. In \cite{arth2015instant}, the authors refine the camera pose by matching the extracted query image’s geometric features and building outlines with a GPS-obtained coarse prior pose. However, GPS signals are difficult to receive indoors, and WiFi, Bluetooth, etc. must be pre-installed and carefully calibrated, both of which create logistical challenges.  In \cite{song20166}, the authors use a VPR algorithm to find a set of similar reference images to the query image and then refine the camera pose with a relative-pose computation algorithm. This algorithm, however, requires use of the camera's intrinsic parameters, which contain the focal length information inherent to the specific camera being used, a step that is difficult to complete in an algorithm that must support multiple end user devices.

\textbf{\textit{The third approach computes the camera pose by reprojecting 3D landmarks in a reconstructed map back to 2D image and minimizing the discrepancies between the observed 2D points and their corresponding reprojections \cite{liu2017efficient,svarm2016city,taira2018inloc,toft2018semantic}.}} Perspective-n-point (PnP) is the most frequently used algorithm to solve this reprojection, computing camera pose using a set of 2D-3D point correspondences between the query image and the reconstructed 3D map. However, it is time-consuming to search for 3D landmark correspondence in the reconstructed map for the 2D features in a query image. To improve computational efficiency, \cite{sarlin2019coarse} introduced a coarse-to-fine strategy that first uses the VPR method to retrieve similar images of the query image and then uses the 3D landmark positions they stored to lessen the search range, enabling precise real-time localization in vast environments.

\section{Method}

In this section, our entire system architecture is first introduced, then we illustrate two types of user interface.

\subsection{System Design and Architecture}

\begin{figure}[H]
\includegraphics[width=\textwidth]{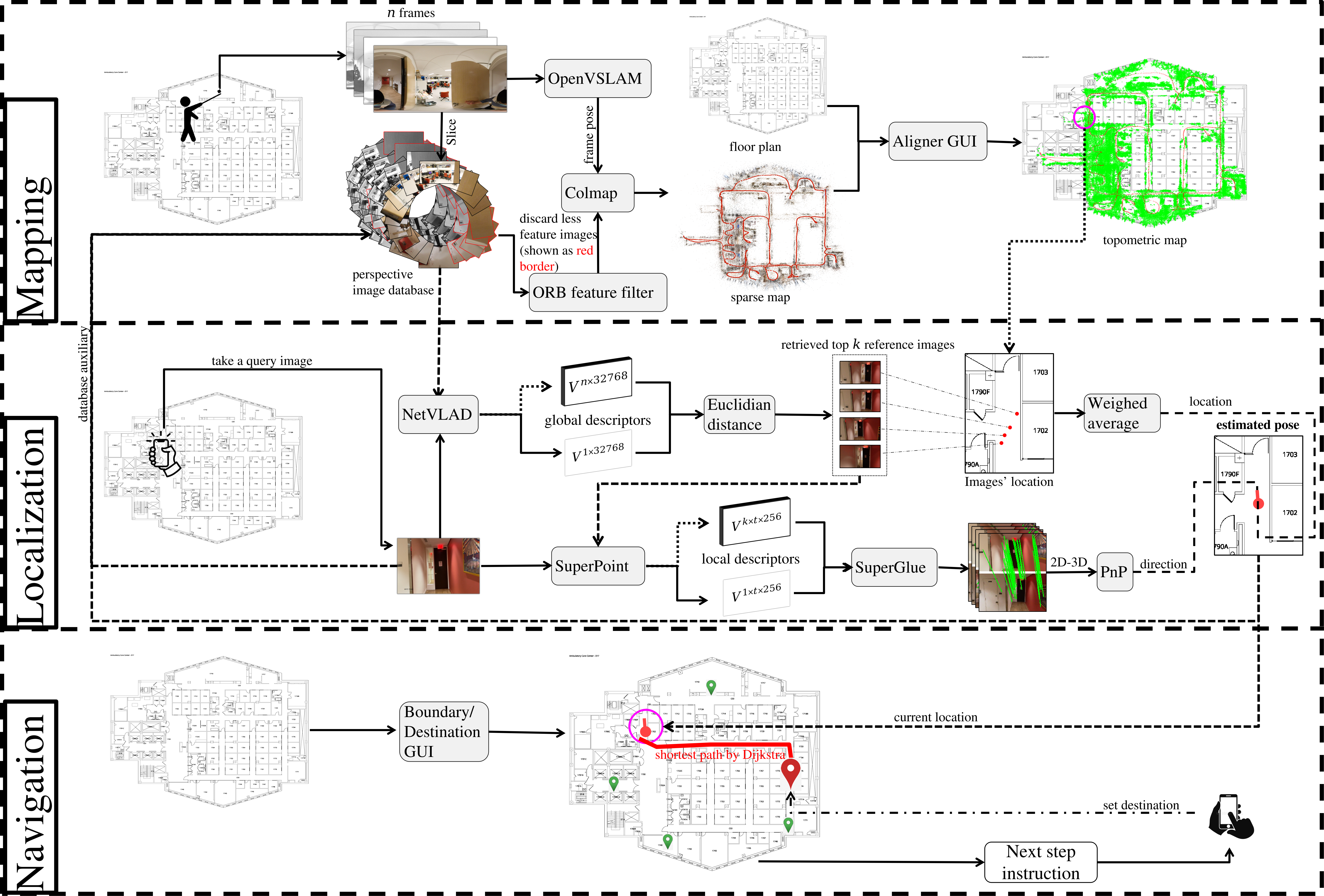}
\caption{\textbf{System architecture diagram. }Our pipeline consists of mapping, localization and navigation module. During the mapping module, OpenVSLAM and Colmap algorithms use 360-degree images as input to generate a topometric map. On the basis of this topometric map, NetVLAD will first retrieve similar images of a query image, followed by two downstream tasks: weight average and coarse-to-fine PnP algorithms that estimate the location and direction of the query image. In the navigation module, a Dijkstra algorithm computes the shortest path between any database images using a navigable graph defined by the topometric map's boundaries. During real-time navigation, our system will utilize both the shortest path information and predefined destinations to guide end-users to their desired destination. The captured query image will be returned to the mapping module to aid in the evolution of the topometric map.\label{fig:1}}
\end{figure}  

Our system can be divided into three phases, mapping, localization, and navigation, as shown in Figure \ref{fig:1}. Using a 360-degree field of view (FOV) camera (to improve the image database creation), a map-maker captures a video of a target space and extracts a sequence of equirectangular frames from this video. These sequential equirectangular images and the corresponding floor plan are fed into the mapping phase to generate a specialized ‘place’ map or the so-called topometric map. This map is then used in the localization and navigation phases. Our system employs a VPR task to retrieve similar images to the query image $I_q$ taken by an end-user, followed by two downstream tasks: weighted averaging and the PnP to estimate the query image’s location and direction. Based on the retrieved location and direction, a shortest path planning algorithm will safely guide the end-user from an origin to a desired destination. We will explain these three phases in detail in the following subsections.

\subsection{Mapping}

The topometric map is generated in this phase, which plays a pivotal role in our entire system. It facilitates the delineation of boundaries around navigable spaces and the identification of destinations that may be of interest to end-users.  Furthermore, it contains a reconstructed 3D sparse map (or raw map) generated from multi-view RGB reference images of the target space and geolocations of these reference images, which are essential for estimating a camera’s location and direction from a query image. To reconstruct this sparse map, one could use simultaneous localization and mapping (SLAM) or structure from motion (SfM) algorithms. The former uses sequential images as input to generate the sparse map in real-time, while the latter uses unordered images and computes the sparse map offline.

OpenVSLAM \cite{sumikura2022openvslam} is a SLAM system based on Orb-slam2 \cite{mur2017orb} that supports multiple camera models, such as the equirectangular camera model, which has a $360$-degree FOV and ensures sufficient overlap between adjacent images that can enhance the robustness of the map reconstruction. It uses ORB features \cite{rublee2011orb} to match two images, which works well when two images are relatively similar but frequently fails when two images have large orientation or position differences. The Superpoint network \cite{detone2018superpoint}, on the other hand, can handle these differences robustly, resulting in a significantly more precise matching result. Colmap \cite{schonberger2016pixelwise,schonberger2016structure}, one of the most popular SfM pipelines, supports Superpoint features. However, it only supports the perspective camera model, which has less than $180$ degrees FOV and therefore cannot guarantee sufficient overlap between adjacent images.

\begin{table}[H] 
\caption{\textbf{Advantages of our methods compared with OpenVSLAM and Colmap.} Our approaches support both the equirectangular camera model and the SuperPoint feature, resulting in strong mapping and localization performance. \label{tab:1}}
\newcolumntype{P}{>{\centering\arraybackslash}X}
\begin{tabularx}{\textwidth}{p{2.4in}p{0.8in}p{0.6in}p{0.8in}}
\toprule
 & \multicolumn{1}{c}{\textbf{OpenVSLAM}} & \multicolumn{1}{c}{\textbf{Colmap}} & \multicolumn{1}{c}{\textbf{Our methods}}\\
\midrule
Support equirectangular camera model\newline (Robust in mapping) & \multicolumn{1}{c}{\checkmark} & \multicolumn{1}{c}{} & \multicolumn{1}{c}{\textbf{\checkmark}}\\
Support SuperPoint feature\newline (Robust in localization) & \multicolumn{1}{c}{} & \multicolumn{1}{c}{\checkmark} & \multicolumn{1}{c}{\textbf{\checkmark}}\\
\bottomrule
\end{tabularx}
\end{table}

To ensure robustness of our system in both mapping and localization, we combine the advantages of OpenVSLAM and Colmap, as listed in Table \ref{tab:1}. We construct a sparse map with OpenVSLAM and enhance it with Colmap by replacing its ORB feature with the SuperPoint feature. The input of our mapping module is a sequence of equirectangular images $I_i\in {\mathbb{R}}^{3840\times 1920\times 3},\ (i=1,2,3,\cdots ,n)$ captured in the target space. Using these images, OpenVSLAM can accurately and robustly reconstruct a sparse map containing each equirectangular image's 3D location $P_i$, direction ${\alpha }_i$, and a set of ORB features. We discard these ORB features and evenly slice $I_i$ into $m=\frac{{360}^{{}^\circ }}{\theta }$ perspective images $I^t_i\in {\mathbb{R}}^{640\times 360\times 3},(t=1,2,\cdots ,m)$ with a width FOV of $\gamma $ degree and a horizontal viewing direction of ${\theta }_t=t\times \theta $, where $\theta $ is the view direction intersection angle between two adjacent perspective images. These perspective images comprise a reference image database which is used in localization and navigation. For each reference image, we extract its SuperPoint features with local descriptor ${d^l}^t_i\in {\mathbb{R}}^{1\times 256}$, compute its direction ${\alpha }^t_i={\alpha }_i+{\theta }_t$, and send ${d^l}^t_i,{\alpha }^t_i$, along with its location $P_i$, into Colmap to reconstruct the sparse map we want. 

\begin{figure}[H]
\begin{adjustwidth}{-\extralength}{0cm}
\centering
\includegraphics[width=1.3\textwidth]{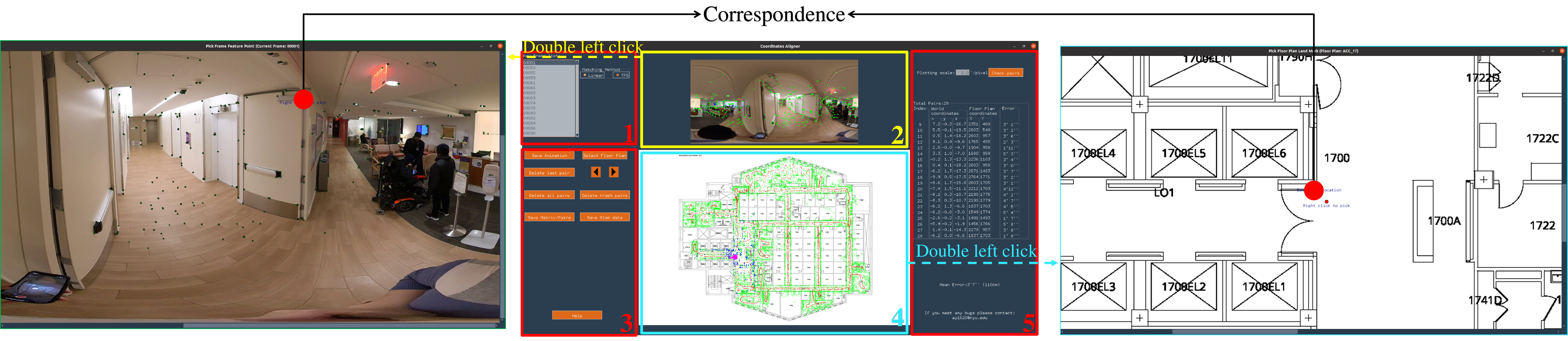}
\end{adjustwidth}
\caption{\textbf{Topometric map aligner GUI.} The main window is displayed in the middle, which comprises five zones. \textit{Zone 1} has a list of all equirectangular images recorded in the target space that can be browsed in \textit{zone 2}. Several buttons in \textit{zone 3} aid the map-maker in projecting the raw map onto the floor plan. The map-maker can upload and display the floor plan of the target space in \textit{zone 4} by clicking the 'Select Floor Plan' button in \textit{zone 3}. Two magnified views (left and right) aid the cartographer in locating the 2D-3D correspondences between the equirectangular image and floor plan. Once the transformation matrix is identified, \textit{zone 5} will display map error information.\label{fig:2}}
\end{figure}  


However, this sparse map is still defined in the 3D coordinate frame in OpenVSLAM (or Colmap), which lacks the necessary boundary information to ensure end-users navigate safely. To solve this problem, we project the sparse map onto a 2D floor plan’s coordinate frame using the transformation parameters between these two coordinate frames and define the relevant boundaries. To compute these transformation parameters, we need to find a set of 2D-3D point correspondences, which can be manually selected from our graphical user interface (GUI), as shown in Figure \ref{fig:2}. When opening this GUI, all equirectangular images captured in the target space are loaded and can be individually selected from the list in \textit{zone 1} for browsing in \textit{zone 2}. Then the map-maker can click the 'Select Floor Plan' button in \textit{zone 3} to upload the target space's floor plan, which will then be displayed in \textit{zone 4}. To facilitate the selection of 2D-3D point correspondences, the map-maker can double left-click in \textit{zone 2} (and \textit{zone 4)} to open a magnified view of the currently selected equirectangular image (and the floor plan). To record a manually identified correspondence (such as two red dots shown in Figure \ref{fig:2}), the map-maker can first click the feature point on the image and then click its corresponding location on the 2D floor plan. Note that each feature point has a 3D coordinate $X_i=(x_i,\ y_i,z_i)$ in the OpenVSLAM (or Colmap), therefore a 2D-3D correspondence is identified. The $y-axis$ of the OpenVSLAM (or Colmap) coordinate frame in our system is perpendicular to the ground plane, and can therefore be neglected from the coordinate transformation; all $y_i$ coordinates are set to $1$. Once the map-maker selects $h\ge 3$ correspondences, we can use Equation (\ref{equ1}) to calculate the transformation matrix,
\begin{linenomath}
\begin{equation}
\label{equ1}
T=xX^T{\left(XX^T\right)}^{-1},
\end{equation}
\end{linenomath}
Here\textit{ }$x:{\mathbb{R}}^{2\times h}$ means a set of 2D floor-plan coordinates, $X:{\mathbb{R}}^{3\times h}$ means the set of corresponding 3D sparse map coordinates, and the resulting transformation matrix $T:{\mathbb{R}}^{2\times 3}$ can convert coordinates from the OpenVSLAM frame to the 2D floorplan frame. Finally, using $T$, the locations of all reference images and the 3D landmark points in the sparse map can be projected onto the floor plan and displayed at\textit{ zone 4 }as red and green dots, respectively. 

\subsection{Localization}

\begin{figure}[H]
\includegraphics[width=\textwidth]{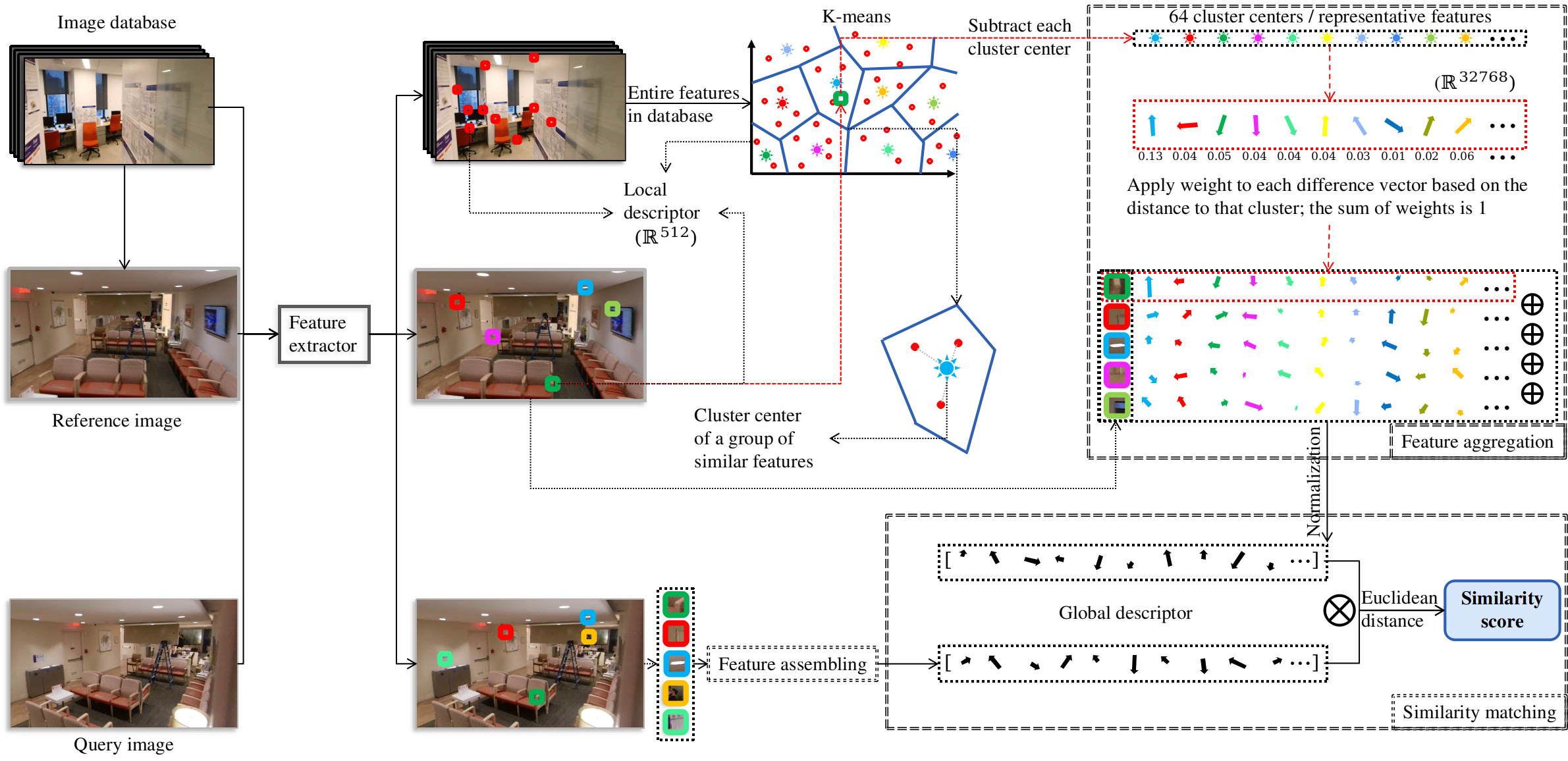}
\caption{\textbf{Retrieval-based localization.} This diagram demonstrates the steps of feature aggregation and similarity matching using NetVLAD. A pretrained NetVLAD extracts several local descriptors of the query image and each reference image, assembling them into global descriptors, and calculating the Euclidean distance between the query image's global descriptor and that of each reference image; the smaller the Euclidean distance is, the more similar the images are.\label{fig:3}}
\end{figure} 

The locations of the reference images and the 3D landmark points are crucial to our end user localization process. In contrast to the method in \cite{song20166} discussed previously, we refine the camera location of the query image $I_q$ by averaging the locations of its top $K$ similar reference images obtained via the VPR task. Similar to \cite{sarlin2019coarse}, we limit the searching range of the 2D-3D correspondences only to these similar reference images to speed up the computation and then use the PnP algorithm on the discovered correspondences to estimate the direction of $I_q$.

To retrieve images that are similar to a given query image from the reference image database, our system first uses NetVLAD to extract the global descriptors  ${d^g}^t_q,,{d^g}^t_i:R^{1\times 32678}$ of $I_q$ and $I^t_i$, then calculates the Euclidean distance between them (Figure \ref{fig:3}) using Equation (\ref{equ2})
\begin{linenomath}
\begin{equation}
\label{equ2} 
D_{qi}=\sqrt{\sum^{32767}_{k=0}{{({d^g}^t_{qk}-{d^g}^t_{ik})}^2}}, 
\end{equation}
\end{linenomath}
The lower the $D_{qi}$ is, the higher similarity score between the reference image $I^t_i$ and the query image $I_q$ is. The reference images with the highest $K$ scores (i.e., the lowest $K$  Euclidean distances) are selected as similar or `candidate' images $I^t_j\ (j=1,2,\cdots K)$. These candidate images are then utilized in two downstream tasks to estimate the end-user's location and direction.

 The first downstream task uses a weighted averaging method to estimate the end-user's location by Equation (\ref{equ3})
\begin{linenomath}
\begin{equation}
\label{equ3}
P=\sum^K_{j=1}{{\omega }_jP_j},
\end{equation}
\end{linenomath}
Here, $P$ is the estimated location of the query image $I_q$. $P_j$ is the location of the candidate image $I^t_j$ on the floor plan, and ${\omega }_j=\frac{m_j}{\sum^K_{k=1}{m_k}}$ is the weight applied on $P_j$, where $m_j$ (or $m_k$) is the number of matched SuperPoint local features between the query image $I_q$ and its candidate image $I^t_j$ (or $I^t_k$) using the SuperGlue network \cite{sarlin2020superglue}. Note that $m_j$ will be set to $0$ if it's not larger than $75$. If $m_j$ of all candidate images are not larger than $75$, the system will set $P$ as the location of the candidate image with the largest $m_j$ that is larger than $30$. If there's no $m_j$ larger than $30$, the system will increase $K$ and retry retrieval until it fails to estimate the camera's location $P$ when $K$ exceeds a threshold. 

\begin{figure}[H]
\includegraphics[width=\textwidth]{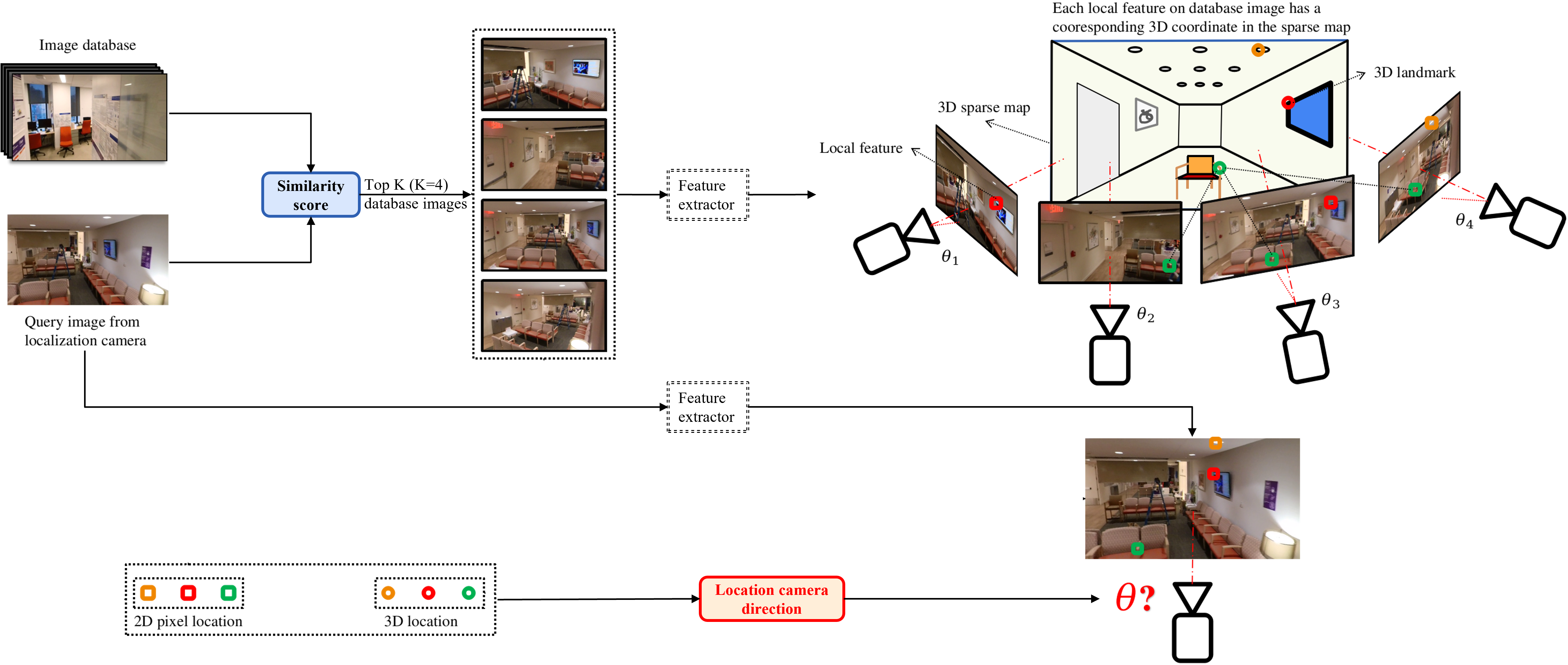}
\caption{\textbf{Hierarchical localization. }Given a query image, the K reference images with the highest matching scores (Figure \ref{fig:3}) are retrieved as candidate images. Using Superpoint and SuperGlue, similar local features between the query image and candidate images can be identified. Each local feature on the candidate image has a 3D location in the sparse map. Thus a set of 2D-3D point correspondence between the query image and the sparse map are found and the direction of the camera is determined using PnP algorithm.\label{fig:4}}
\end{figure} 

The second downstream task efficiently estimates the camera's direction using a coarse-to-fine strategy \cite{sarlin2019coarse}. Specifically, the candidate image $I^t_j$ stores the 3D location of each of its 2D SuperPoint local features in the sparse map, after matching $m^q_j$ SuperPoint local features between $I_q$ and $I^t_j$ using the SuperGlue network. We are therefore able to obtain $\sum^K_{j=1}{m_j}$ 2D-3D point correspondences between $I_q$ and the sparse map, allowing us to efficiently calculate the camera's direction using PnP algorithm (Figure \ref{fig:4}). 

\subsection{Navigation}

\begin{figure}[H]
\begin{adjustwidth}{-\extralength}{0cm}
\centering
\includegraphics[width=1.3\textwidth]{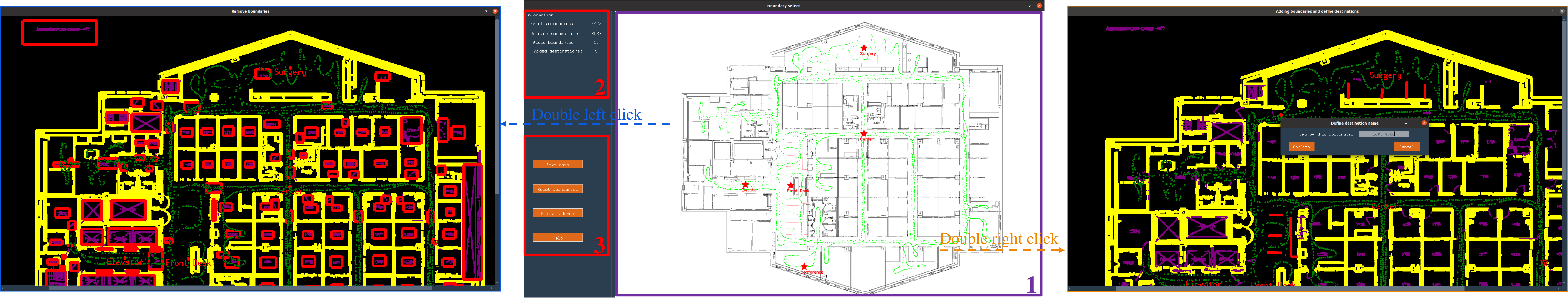}
\end{adjustwidth}
\caption{\textbf{Boundary/Destination GUI.} The main window is displayed in the \textit{middle}. It comprises three zones. The green dots in \textit{zone 1} depict the locations of database images, while the red star indicates potential destinations. After double-clicking \textit{zone 1}, map-makers can adjust boundaries and desired destinations in a additional pop-up windows (\textit{left} and \textit{right
}). After double left-click \textit{zone 1}, the map-maker can delete boundaries in the pop-up window (\textit{left}) by removing line segments (which represent boundaries) using an edit tool (red superimposed box). After double right-click \textit{zone 1}, another pop-up window (\textit{right}) will appear, in which the map-maker can draw red lines to represent additional boundaries, and define potential destinations by clicking green dots and assigning destination name to them. Additionally, boundaries and destination information will be provided in the \textit{zone 2 (middle)}. Once all the boundaries and destinations are defined, the map can be saved in \textit{zone 3}.\label{fig:5}}
\end{figure}  

After retrieving the current location and direction, the navigation module will guide the end-user to the desired destination. A good navigation module should provide the end-user with up-to-date boundary information for safe travel, as well as flexible and abundant destination options. We developed a GUI as depicted in Figure \ref{fig:5} to define boundaries and destinations. It extracts all line segments from the floor plan image to represent potential boundaries and displays them on the topometric map in \textit{zone 1} as in Figure \ref{fig:5}. However, some boundaries might differ from the real world due to the quality of the floor plan or changes in the scene, requiring manual addition or deletion of boundaries in an interactive fashion. This GUI enables map-makers to maintain the map by removing or adding boundaries on the topometric map and redefining desired destinations quickly and efficiently. The left and the right areas shown in Figure \ref{fig:5} are magnified views of the floor plan displayed in \textit{zone 1}.The map-maker can remove boundaries in (Figure \ref{fig:5}, \textit{left}) when \textit{zone 1} is double left-clicked, or add boundaries or define destinations of interest in (Figure \ref{fig:5}, \textit{right}) when \textit{zone 1} is double right-clicked. Each green dot in \textit{zone 1} indicates a reference image in the database. To define a desired destination, the mapmaker must select any one of the reference images in the topometric map that is adjacent to an area of interest and assign it a destination name (Figure \ref{fig:5}, \textit{right}). Note that in our future work, we could utilize object/text detection methods to automatically detect each room's number during the video capture and assign a destination to a reference image frame near that room.

Using the locations of the reference images as the potential destinations has accuracy and safety benefits. Because our localization method is based on VPR, which find similar database images of a query image, so our localization will become more and more accurate as the end-user moves closer to the destination that is defined using the location of a database image, which increases the probability of successfully retrieving similar images to the query image. In addition, the reference images were captured by the map-maker, indicating that the area surrounding these reference images is navigable, thereby guaranteeing the safety of the BLV.

Due to the accuracy and safety benefits provided by the reference images, our system is designed to navigate the end-user as closely as possible to the reference images. To accomplish this, it first determines if there exists an immediately navigable path between any pair of reference images by checking if there are boundaries between them. The paths calculated from all image pairs constitute a navigable graph, and the Dijkstra algorithm is used to compute the shortest path between any pair of images based on this graph. This computation can be done off-line, and the information can be quickly updated if the boundaries change. During the real-time navigation, the end-user is required to select a desired destination from the destination list defined by the map-maker. Once the end-user has been localized via a query image, the system will first direct the user to the closest reference image's location, and then direct them along the shortest route to reach the destination.

\subsection{User Interface}

\begin{figure}[H]
\includegraphics[width=\textwidth]{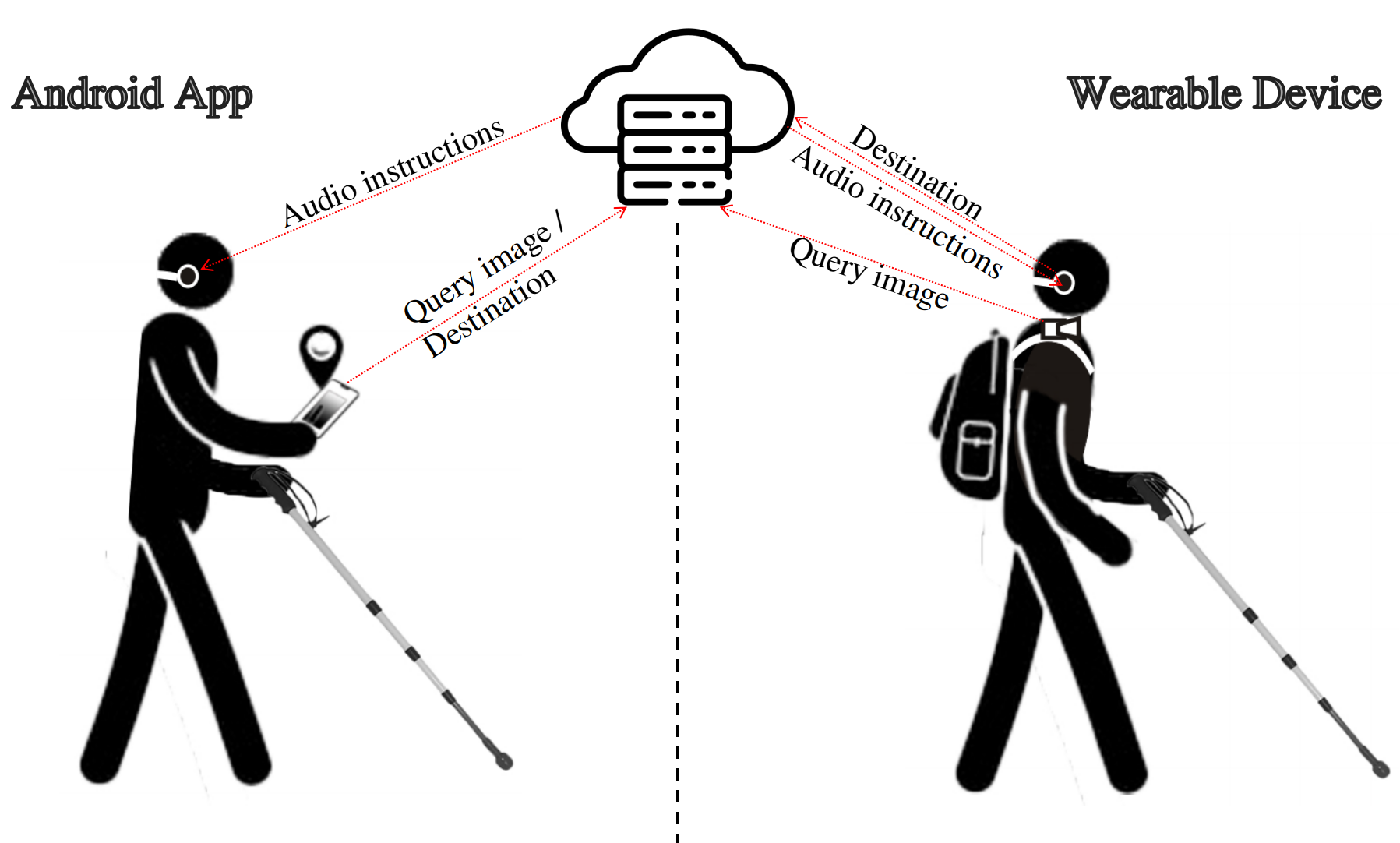}
\caption{\textbf{Two types of end-user interfaces.} On the left is an infographic that demonstrates the use of our Android application. The user selects the desired destination and sends it to the server. For localization, the end-user touches the screen to capture a query image and it is sent to the cloud server. The server then calculates the camera pose and sends instructions to the end-user via a binaural bone-conduction headset. On the right is an infographic that demonstrates the use of our wearable device with a binaural bone-conduction headset. During navigation, a camera connected to a backpack-mounted NVIDIA {\circledR} Jetson AGX Xavier captures and sends query images to a cloud server; the user receives navigation instructions via the headset.\label{fig:6}}
\end{figure} 

The end-user can navigate using either of the two user interfaces we designed (Figure \ref{fig:6}). One is for an Android application installed on the Android device, and another is for a wearable device, which employs a discreet USB camera tethered to a micro-computer housed in a backpack.

\subsubsection{Android Application}

The Android application contains a navigation bar, as shown in Figure \ref{fig:7}. The end-user needs to select the current building, floor, and the desired destination on their cell phone. Once the destination is selected, the phone’s camera will activate; the end-user will need to hold the phone in landscape view and tap the screen to capture a query image which will then be sent automatically to a server to calculate current location and direction, after which a navigation prompt will be delivered. Our system supports another automated camera acquisition mode that intermittently takes the query image every predetermined number of seconds without requiring the end-user to tap the screen. Note that the phone requires access to the camera, and it can either be manually held (less preferred) or simply positioned in a lanyard at chest-level (more preferred). All touch-based operations in this application can be replaced with speech prompts to reduce operational difficulties for the BLV. 

\begin{figure}[H]
\includegraphics[width=\textwidth]{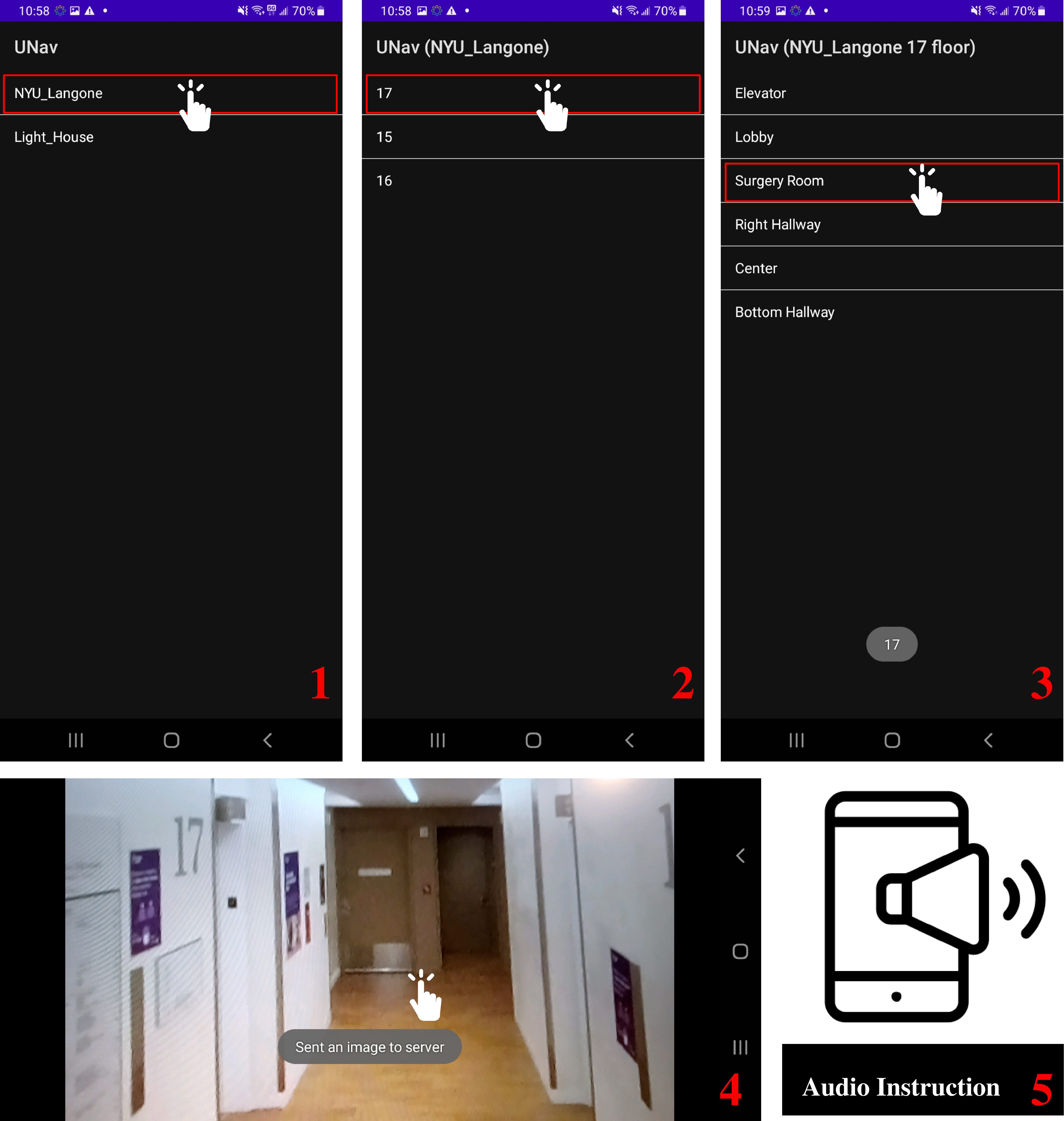}
\caption{\textbf{Android application.} There are five steps to any trip initiation. The first three steps are a cascade of linked choices that must be completed for the user to select the desired destination, inclusive of building, floor and room. This procedure may be completed with either verbal or tactile input. Once the destination has been selected, the camera will turn on and the user will need to touch the screen to take and send an initial query image to the server (step 4). The server will then send back navigation instructions (step 5). After multiple iterations of steps 4 and 5, the user will reach the desired destination.\label{fig:7}}
\end{figure} 

\subsubsection{Wearable Device}

Cell phones are ubiquitous but pose challenges when used for sustained periods, particularly when the camera feed is being used intermittently. In order to address the ergonomics of this problem and to improve image quality, we have developed a backpack \cite{rizzo2017somatosensory,niu2017wearable,shoureshi2017smart} with an NVIDIA ® Jetson AGX Xavier connected to a battery, USB camera, and a binaural bone-conduction headset (Figure \ref{fig:6}, \textit{right}). The battery supplies power for the hardware; the USB camera is used to take query images. The end-user can send vocal commands through the microphone in the headset and receive audio prompts from the server. 
\section{Evaluation}
\subsection{Overview}
In this section, we present experimental evaluations of the developed system. There were $6$ participants ($4$ male and $2$ female, with an average age of $32$), including two end-users with lived experience from blindness (one congenitally blind and another with a degenerative retinal dystrophy), who participated in the evaluation process of the developed system (Figure \ref{fig:8}). 

\begin{figure}[H]
\includegraphics[width=\textwidth]{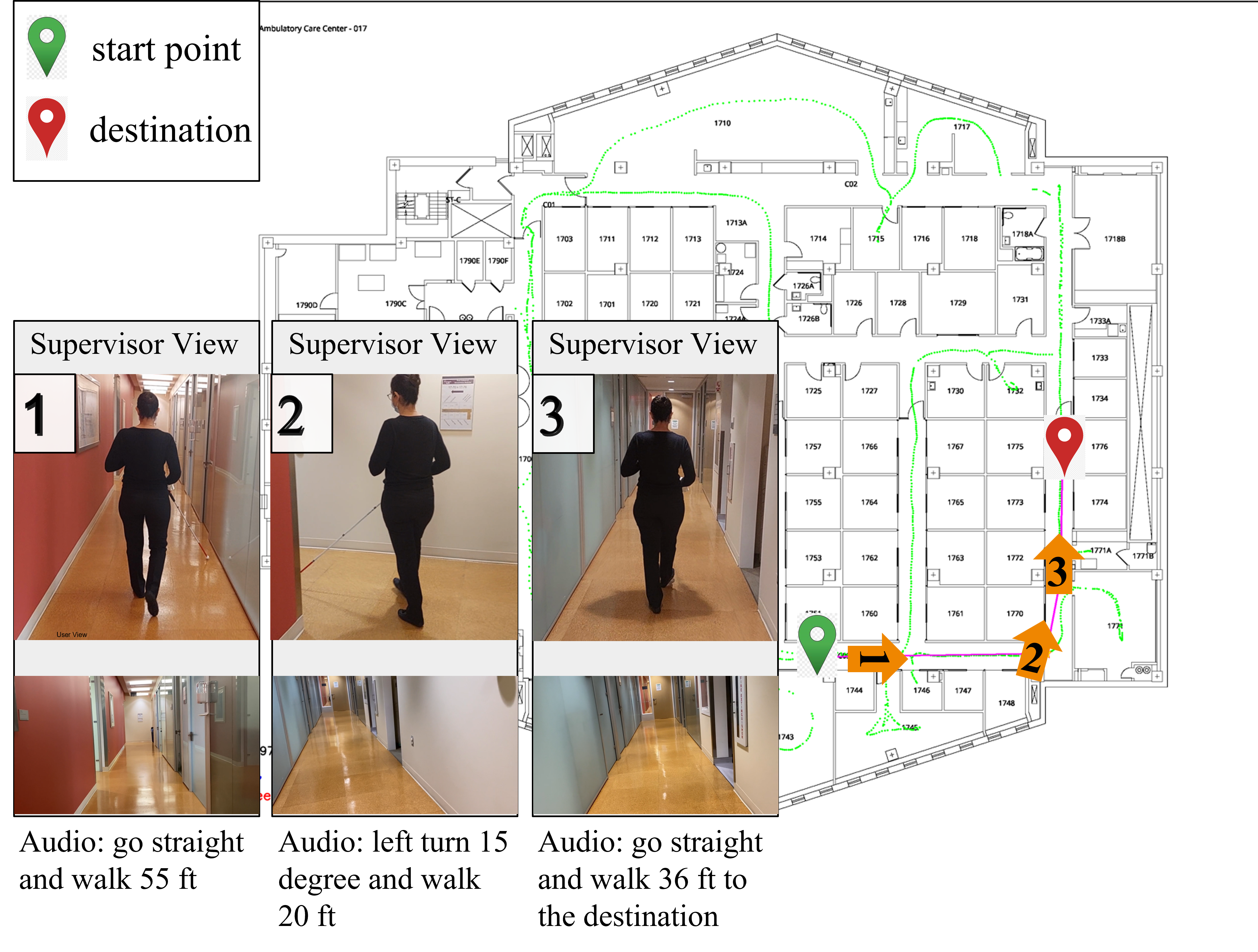}
\caption{\textbf{Navigation example.} Using our Android application, a member of our team navigates from the origin to the destination safely.\label{fig:8}}
\end{figure} 

\subsection{Dataset}

The evaluation was performed at an academic medical center in an ambulatory division within NYU Langone Health (\textit{New York University Langone Ambulatory Care Center, USA}).

A map-maker on our team used an Insta360 camera to collect equirectangular videos along a pre-designed `zigzag' trajectory to ensure the reference image database included maximal features from the target space. This trajectory included $3$ loops in the target space. The first loop included the main hallway with all doors opened, the second loop included all hallways and the entrance into each room with their  doors opened, and the third loop included the whole space with all doors closed (meaning opened by the videographer during mapping). This process was designed through a trial-and-error process. We found the best camera height for pano-videos was approximately 6ft, given the distance between camera and ceiling in this particular space, attempting to capitalize on an aerial perspective while being mindful of proximity to the ceiling. After extracting whole equirectangular frames ($n=4258$) from the video, we evenly sliced each of them into $m=18$ perspective images with $\theta $=${20}^{{}^\circ }$. Each perspective image has a size of $640\times 360$ and width FOV of $\gamma ={75}^{{}^\circ }$. These $18$ images were filtered to avoid perceptual aliasing by counting the valid feature points extracted by the ORB detector; images were removed if the amount of valid feature points fell below $100$.

\subsection{Localization Evaluation}

Localization accuracy underpins navigation accuracy. Thus, we evaluated the localization accuracy of our system.

To test the system's overall localization accuracy, we first selected $17$ points on the floor plan as testing locations, which corresponds to locations that are easily identified in the real world, such as the corner of structural columns and doorframes, and measured their pixel coordinates on the floor plan, as ground truth locations. Each participant captured testing images at each testing location in the real environment with a ground truth direction obtained by a compass. The location/direction error is computed based on the Euclidean distance/absolute difference between the ground truth location/direction and the estimated location/direction. To draw a convincing conclusion, we averaged the error computed by all participants.

Since our system uses an image retrieval method, there is a natural hypothesis that the denser the reference image database is, the more accurate the localization that can be achieved. To test this hypothesis, we designed two downsampling experiments based on the delineated dataset:  

\begin{itemize}
\item  \textbf{\textit{Frame downsampling}.} We evenly downsample the $n=4258$ equirectangular frames with a downsampling rate $\alpha \in \left[1,5,10,15,20,25,30,40,50\right]$ and slice them into $m=18$ perspective images to form a reference image database.

\item  \textbf{\textit{Direction downsampling}.} We maintain the original number of equirectangular frames equal to $n$. Then after slicing each frame into $m=18$ perspective images and filtering into $\hat{m}$ valid slices, we evenly downsample the slices with a downsampling rate $\beta \in \left[1,2,\cdots ,6\right]$ to form a reference image database.
\end{itemize}
\begin{figure}[H]
\includegraphics[width=\textwidth]{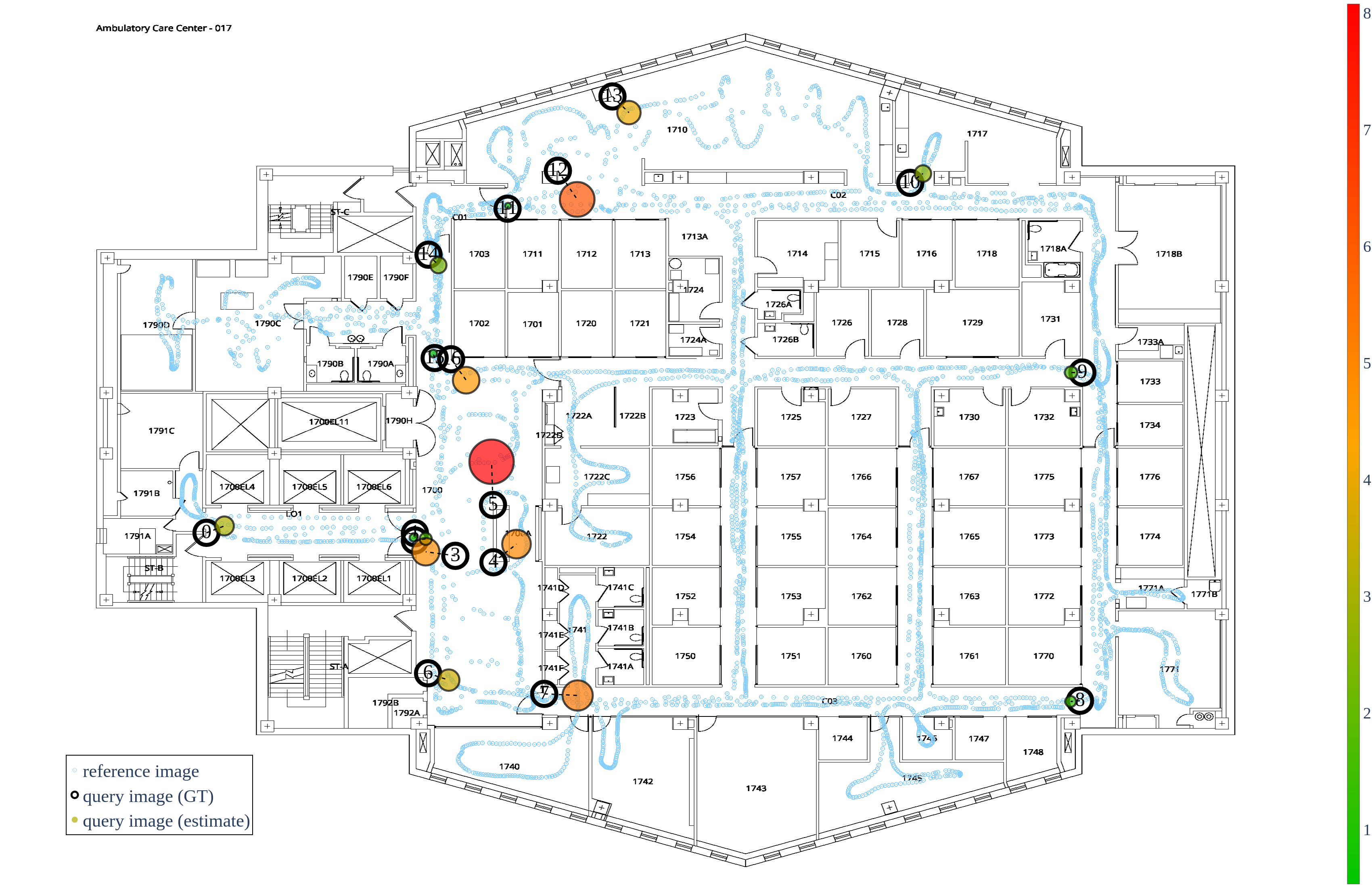}
\caption{\textbf{ Location accuracy heatmap.} The small open light-blue dots are the locations of $n$ reference images on the floor plan. The open black circles are the ground truth locations $17$ testing images; the multi-colored circles next to the GT localizations colored from green to red are the estimated locations of these $17$ testing images; the further from the ground truth location (ft), the bigger the circle and the warmer the color (red).\label{fig:9}}
\end{figure} 

\section{Test Results}

In this section, we calculate the location and direction errors at each of the $17$ test locations based on the two experiments.

\subsection{Localization Results}

We visualize the localization error on a heat map (green indicates less error; red indicates more error) of our target space; each testing location is demarcated with an open circle, as shown in Figure \ref{fig:9}. Here, we set both the $\alpha$ and $\beta$ to $1$ which means no downsampling operations on the original dataset. Under this configuration, the image database is the densest, but there is still a wide variation of location errors across 17 testing locations. The reason behind this phenomenon is that even though the reference image database (represented by the blue dots) is the densest, it remains challenging for map-makers to cover the entire floor plan when recording the reference video. Thus, when determining the location of a query image by applying the weighted average to the geolocations of its candidate reference images, the error will be large if the query image was captured in areas with insufficient reference images. To facilitate the analysis of the correlation between map density and estimated location precision, two tables based on the two evaluation settings are provided below. In these two tables, we examine the systematic decline in localization accuracy as a result of data downsampling.

\begin{table}[H] 
\caption{Estimated location errors (ft) at \textit{$17$} testing locations with different frame downsampling rate \textit{$\alpha \in \left[1,5,10,15,20,25,30,40,50\right]$} on \textit{$n=4258$} reference images. \label{tab:2}}
\newcolumntype{P}{>{\centering\arraybackslash}X}
\begin{tabular}{|p{0.1in}|p{0.6in}|p{0.3in}p{0.3in}p{0.3in}p{0.3in}p{0.3in}p{0.3in}p{0.3in}p{0.3in}p{0.3in}|} \hline 
\multicolumn{11}{|c|}{Different Frame Sampling Density} \\ \hline 
 & Error (ft) & $n/1$ & $n/5$&$n/10$&$n/15$&$n/20$&$n/25$&$n/30$&$n/40$&$n/50$\\ \cline{2-11}
 \parbox[t]{3.3in}{\multirow{17}{*}{\rotatebox[origin=c]{90}{Different Testing Locations}}} & \multicolumn{1}{c|}{$0$} & $3.1$&$3.1$&$2.5$&$2.4$&$2.6$&$2.6$&$2.7$&$3.1$&$3.8$\\ 
 & \multicolumn{1}{c|}{$1$} & $0.9$&$1.6$&$1.6$&$1.7$&$1.5$&$2.4$&$1.1$&$1.4$&$1.4$\\ 
 & \multicolumn{1}{c|}{$2$} & $1.8$&$2.3$&$2.1$&$3.7$&$3.2$&$9.2$&$3.4$&$2.0$&$7.3$\\ 
 & \multicolumn{1}{c|}{$3$} & $4.8$&$4.2$&$3.1$&$4.9$&$1.2$&$5.1$&$4.6$&$5.1$&$5.7$ \\ 
&\multicolumn{1}{c|}{$4$} & $5.0$&$3.3$&$4.3$&$5.9$&$3.4$&$7.2$&$5.9$&$11.1$&$7.1$ \\ 
& \multicolumn{1}{c|}{$5$} & $8.1$&$9.1$&$10.7$&$5.3$&$10.4$&$8.1$&$10.7$&$4.9$&$8.6$ \\ 
& \multicolumn{1}{c|}{$6$} & $3.6$&$4.3$&$4.5$&$4.1$&$5.3$&$4.2$&$14.4$&$4.7$&$20.2$ \\ 
& \multicolumn{1}{c|}{$7$} & $5.4$&$5.4$&$5.4$&$5.4$&$11.8$&$53.0$&$6.0$&$8.5$&$6.5$ \\ 
& \multicolumn{1}{c|}{$8$} & $1.5$&$3.0$&$1.8$&$5.5$&$17.4$&$2.7$&$17.3$&$17.1$&$14.2$ \\
& \multicolumn{1}{c|}{$9$} & $1.8$&$3.3$&$3.3$&$1.6$&$24.6$&$4.1$&$22.9$&$22.3$&$18.9$ \\
& \multicolumn{1}{c|}{$10$} & $2.7$&$1.9$&$3.0$&$3.0$&$3.2$&$3.0$&$3.0$&$3.2$&$3.0$ \\
& \multicolumn{1}{c|}{$11$} & $0.5$&$0.7$&$0.7$&$4.6$&$0.7$&$4.8$&$53.2$&$0.7$&$11.6$ \\ 
& \multicolumn{1}{c|}{$12$} & $6.2$&$3.7$&$3.7$&$8.3$&$8.6$&$10.2$&$8.3$&$8.6$&$10.2$ \\ 
& \multicolumn{1}{c|}{$13$} & $4.0$&$17.7$&$5.4$&$8.4$&$5.8$&$5.0$&$4.6$&$3.4$&$8.0$ \\ 
& \multicolumn{1}{c|}{$14$} & $2.5$&$2.3$&$3.1$&$3.7$&$2.8$&$1.9$&$3.7$&$12.7$&$4.5$ \\  
& \multicolumn{1}{c|}{$15$} & $0.9$&$1.2$&$2.5$&$10.1$&$2.5$&$14.2$&$45.6$&$2.5$&$12.1$ \\
& \multicolumn{1}{c|}{$16$} & $4.6$& $3.7$& $5.4$& $5.1$& $6.2$& $17.0$& $5.8$& $4.8$& $20.1$\\ \hline 
\end{tabular}
\end{table}

Table \ref{tab:2} displays the estimated location errors at $17$ testing locations with different downsampling rates $\alpha $ on the $n$ reference images. Just by reading this table, it is difficult to determine whether this downsampling compromises the estimated location precision. Thus, we utilize Equation (\ref{equ4})
\begin{linenomath}
\begin{equation}
\label{equ4}
p=\frac{1}{17\times 9\times 4}\sum^{16}_{i=0}{\sum^8_{j=0}{\sum^8_{k=j+1}{I_A\left(E_{ij},E_{ik}\right)}}},
\end{equation}
\end{linenomath}
Here $I_A\left(E_{ij},E_{ik}\right)=\left\{ \begin{array}{c}
1,\ \ if\ E_{ij}\le E_{ik} \\ 
0,\ \ \ \ otherwise \end{array}
\right.$, where $E_{ij},\ E_{ik}$ are cells in row $i$, column $j,k$. This function computes the probability, denoted by a variable $p$, that a cell in Table \ref{tab:2} is not greater than any cell to its right in the same row. It iterates through each row and compares each cell to those to its right, counting 1 if the value on the left is not greater than that on the right, i.e., smaller errors on the left and bigger errors on the right. After this iteration and normalization of the counted number, we obtained the probability $p=0.72$, which is greater than $0.5$, indicating that under the frame downsampling setting, the location estimation error by using a denser map (a value on the left) is indeed generally smaller than that of using a sparser map (a value on the right).

However, this frame downsampling setting is insufficient to prove our hypothesis because it has little effect on the direction variance of the reference image database, which is crucial for image retrieval. Consequently, we applied the second direction downsampling setting to see if direction downsampling also compromises the estimated location precision. Table \ref{tab:3} displays the estimated location errors at $17$ testing locations with different downsampling rates $\beta$ on the $m$ perspective images of each equirectangular frame. Using a function similar to Equation (\ref{equ4}), we get $p=0.65$, which is also greater than $0.5$, indicating that the direction downsampling also reduces the accuracy of the location estimation.

\begin{table}[H] 
\caption{Estimated location errors (ft) at \textit{$17$} testing locations with different direction downsampling rate \textit{$\beta \in [1,2,\cdots ,6]$} on \textit{$\hat{m}\le 18$} filtered perspective images of each equirectangular image. \label{tab:3}}
\newcolumntype{P}{>{\centering\arraybackslash}X}
\begin{tabular}{|p{0.1in}|p{0.6in}|p{0.55in}p{0.55in}p{0.55in}p{0.55in}p{0.55in}p{0.55in}|} \hline 
\multicolumn{8}{|c|}{Different Frame Sampling Density} \\ \hline 
 & Error (ft) & $m/1$&$m/2$&$m/3$&$m/4$&$m/5$&$m/6$\\ \cline{2-8}
 \parbox[t]{3.3in}{\multirow{17}{*}{\rotatebox[origin=c]{90}{Different Testing Locations}}} & \multicolumn{1}{c|}{$0$} & $3.1$&$2.8$&$2.8$&$3.0$&$4.1$&$4.5$\\ 
 & \multicolumn{1}{c|}{$1$} & $0.9$&$0.9$&$1.8$&$3.0$&$3.8$&$3.8$\\ 
 & \multicolumn{1}{c|}{$2$} & $1.8$&$3.4$&$3.4$&$3.8$&$7.1$&$2.7$\\ 
 & \multicolumn{1}{c|}{$3$} & $4.8$&$5.3$&$4.0$&$2.6$&$4.2$&$5.2$ \\ 
&\multicolumn{1}{c|}{$4$} & $5.0$&$5.4$&$5.0$&$7.5$&$1.3$&$8.0$ \\ 
& \multicolumn{1}{c|}{$5$} & $8.1$&$7.9$&$8.1$&$9.3$&$8.1$&$9.1$ \\ 
& \multicolumn{1}{c|}{$6$} & $3.6$&$4.1$&$5.3$&$4.0$&$7.9$&$1.5$ \\ 
& \multicolumn{1}{c|}{$7$} & $5.4$&$2.1$&$5.4$&$5.4$&$5.4$&$112.3$ \\ 
& \multicolumn{1}{c|}{$8$} & $1.5$&$2.1$&$1.8$&$1.7$&$1.4$&$1.7$ \\
& \multicolumn{1}{c|}{$9$} & $1.8$&$2.1$&$1.8$&$2.1$&$1.1$&$1.1$ \\
& \multicolumn{1}{c|}{$10$} & $2.7$&$2.7$&$2.1$&$2.7$&$2.1$&$2.1$ \\
& \multicolumn{1}{c|}{$11$} & $0.5$&$0.5$&$0.7$&$3.0$&$0.6$&$39.6$ \\ 
& \multicolumn{1}{c|}{$12$} & $6.2$&$4.3$&$4.6$&$0.8$&$7.3$&$5.8$ \\ 
& \multicolumn{1}{c|}{$13$} & $4.0$&$3.9$&$3.7$&$17.7$&$17.7$&$15.9$ \\ 
& \multicolumn{1}{c|}{$14$} & $2.5$&$3.1$&$2.9$&$2.7$&$2.5$&$1.8$ \\  
& \multicolumn{1}{c|}{$15$} & $0.9$&$0.9$&$0.9$&$0.9$&$0.9$&$0.9$ \\
& \multicolumn{1}{c|}{$16$} & $4.6$& $5.8$& $5.4$& $5.1$& $5.7$& $5.2$\\ \hline 
\end{tabular}
\end{table}

\subsection{Direction Results}

These data so far partially support our hypothesis that the denser the reference image database is in our system, the more accurate our location estimation is. However, we still need to determine whether frame or direction downsampling compromises the accuracy of the direction estimation. Table \ref{tab:4} shows the mean direction estimation errors for different frame/direction downsampling settings.

In contrast to the location estimation error, this table indicates that the overall direction estimation error is negligible. This is because that PnP algorithm leverages 2D-3D point correspondences between the query image and the sparse map, which is less dependent on the density of the reference image database, and therefore more robust. However, the PnP algorithm fails on a few testing points when $\alpha $ or $\beta $ becomes large (especially for $\beta $). This is because when reference images are too sparse, it is difficult for the system to find sufficient 2D-3D point correspondences because few or no candidate images have overlap views with the query image. 

\begin{table}[H] 
\caption{Mean estimated direction errors (\textit{${}^\circ $}) of \textit{$17$} testing locations under different combination of frame downsampling rate \textit{$\alpha $} and direction downsampling rate \textit{$\beta $}. \textit{$n/s$} indicates that directions at some testing locations cannot be retrieved. \label{tab:4}}
\newcolumntype{P}{>{\centering\arraybackslash}X}
\begin{tabular}{|p{0.4in}|p{0.6in}|p{0.5in}p{0.5in}p{0.5in}p{0.5in}p{0.5in}p{0.5in}|} \hline 
\multicolumn{8}{|c|}{Different Direction Sampling Density} \\ \hline 
 & Error (ft) & $m/1$&$m/2$&$m/3$&$m/4$&$m/5$&$m/6$\\ \cline{2-8}
 \parbox[t]{3.3in}{\multirow{9}{*}{\rotatebox[origin=c]{90}{Different Frame Sampling Density}}} & \multicolumn{1}{c|}{$n/1$} & $2$&$3$&$3$&$2$&$2$&$13$\\[6pt] 
 & \multicolumn{1}{c|}{$n/5$} & $2$&$2$&$2$&$2$&$3$&$3$\\[6pt]  
 & \multicolumn{1}{c|}{$n/10$} & $2$&$2$&$2$&$2$&$2$&$2$\\[6pt]  
 & \multicolumn{1}{c|}{$n/15$} & $3$&$3$&$n/s$&$3$&$3$&$n/s$ \\[6pt]  
&\multicolumn{1}{c|}{$n/20$} & $2$&$3$&$2$&$3$&$3$&$3$ \\[6pt]  
& \multicolumn{1}{c|}{$n/25$} & $4$&$3$&$3$&$4$&$4$&$14$ \\[6pt]  
& \multicolumn{1}{c|}{$n/30$} & $3$&$3$&$n/s$&$3$&$3$&$n/s$ \\[6pt]  
& \multicolumn{1}{c|}{$n/40$} & $2$&$3$&$2$&$3$&$2$&$n/s$ \\[6pt]  
& \multicolumn{1}{c|}{$n/50$} & $4$&$3$&$n/s$&$n/s$&$4$&$n/s$ \\[6pt] 
\hline 
\end{tabular}
\end{table}

\section{Discussion}

\subsection{Technical Underpinnings of Navigation Solutions}

Broadly speaking, navigation methods for assistive technologies can be categorized as sensor-based and vision-based. The localization accuracy for most sensor-based systems suffers from accuracy and reliability concerns. Although a handful of sensor-based navigation solutions are able to track location precisely and robustly, most require pre-installed and carefully calibrated physical sensor infrastructure, which is costly, time-consuming, or often infeasible at-scale. To overcome these obstacles, our system employs a vision-based localization system that requires only ordinarily utilized cameras for data capture and can provide comparable accuracy on the location and direction estimation. 

Moreover, most of the sensor-based systems are difficult to deploy in outdoor environments at a large scale and therefore create handoff problems when wayfinding involves both indoor and outdoor environments. Many existing indoor navigation systems utilize Ultra-wideband (UWB), which requires careful sensing infrastructure installation and calibration \cite{denis2004nlos}. Other indoor technologies, such as Wi-Fi, Bluetooth Low Energy (BLE), and Chirp Spread Spectrum, also require preinstalled infrastructure with a power source; almost all are impractical to deploy in larger, complicated outdoor spaces. Additionally, The vast majority of existing outdoor navigation systems rely on GPS signal, which is difficult to receive in indoor environments, and when used outdoors, suffers from larger errors due to multi-path or 'urban canyon' \cite{hsu2015evaluation} Since our system uses only ordinarily utilized cameras for data capture, it is easily deployable in both indoor and outdoor environments and smooths transitions in between indoor/outdoor spaces, obviating the need to translate approaches from one sensor to another.

Compared to the sensor-based systems, most vision-based systems offer a robust path forward, but there are still limitations. Many vision-based approaches require intrinsic camera parameters. Obtaining these parameters presents logistical difficulties, especially for individuals with BLV. Without such information, location estimation errors are frequently very large. Herein, we employed a novel approach with a weighted average algorithm to solve this challenge; it can begin working accurately with a sparse map, and, as the user continues to employ the system and thus increases the map’s density, the map evolves and the localization accuracy continuously improves.

\subsection{Pratical Implications}

Our system is underpinned by video recordings that take on average $30/10.000$ (minutes/sq feet), and generates a respective topometric map with registration between 2D and 3D in approximately $15$ minutes, significantly less time than competing sensor-based solutions that require manual annotation. Moreover, our system can work jointly with janitorial (cleaning) robots and other citizen science opportunities to collect the relevant data required to generate the maps a priori. The boundaries / destinations GUI enables the map-maker to easily update information regarding map boundaries and destinations, allowing the system to rapidly adapt to changing environments, which is difficult for other vision-based and sensor-based approaches. In addition, our system can function without a cell signal by moving all computation onto the edge device; in other words, a relevant map of interest can be downloaded in advance, and the audio instructions can help people with BLV safely reach their destination of interest. Our system achieves positional and directional errors of $1$ meter and $2$ degrees, respectively, according to our evaluation, a considerable advance over other vision-based methods \cite{Kendall_2017_CVPR}.

\subsection{Limitations and Future Directions}

We do anticipate that our system will require a considerable amount of time to exhaustively searching for similar reference images when databases grow to the size of a city; consequently, we could replace the existing method with a more advanced searching algorithm, such as KD-tree, to improve localization efficiency. In addition, even though the evaluation demonstrates that our system could achieve accurate localization with an average error of less than $1$ meter in a large indoor space, if the database is sufficiently dense, we believe we can reduce the localization error even further if we can estimate the camera direction without intrinsic parameters. In \cite{pan2022camera}, the author presents an implicit distortion model that enables optimization of the 6-degree-of-freedom camera pose without explicitly knowing intrinsic parameters. In our future work, we will integrate this method, perform additional evaluations and consider future pipeline upgrades. Finally, our system has been evaluated in an indoor environment; however, its performance in an outdoor environment has not yet been determined. It is difficult to obtain an accurate floor plan for the outdoor spaces. \cite{chen2022heat} propose an attention-based neural network for structured reconstruction for use in outdoor environments. The pipeline takes a 2D raster image as input and reconstructs a planar graph representing the underlying geometric structure; this new approach may afford us the ability to use a satellite images to generate floor plans for use in our future work.
\section{Conclusion}

Herein, a prototype vision-based localization system has been introduced. This system does not require any pre-installed sensor infrastructure or a camera’s intrinsic matrix. At present, the system uses a 360 camera to collect the initial reference image database in the mapping phase, then simple cell phones are used to that acquire additional image frames from multiple vantage points and create denser maps. The localization phase of the system only requires a daily-use camera, as found in most smart phones and tablets. The system is fashioned into a mobile application that can be downloaded on any smart device equipped with a camera and internet connection or onto ergonomic wearables, as illustrated by our novel backpack embodiment. Our goal for this approach is to support navigation, of short and long length, in both indoor and outdoor environments, with seamless handoffs. In the future, such a system could support additional microservices, such as obstacle avoidance or drop-off detection, evolving state-of-the-art wayfinding to a more integrated approach that blends orientation with travel support.
\endgroup
\vspace{6pt} 


\authorcontributions{Conceptualization, C.F. , JR.R. , and A.Y.; methodology, A.Y. , C.F. , JR.R; software, A.Y.; validation, A.Y.; writing—original draft preparation, A.Y.; writing—review and
editing, JR.R. , C.F. , T.H. , R.V. , W.R. , P.M. and M.B. . All authors have read and agreed to the published version of the manuscript.}

\funding{Research reported in this publication was supported in part by the NSF grant 1952180 under the Smart and Connected Community program, as well as by NSF Grant ECCS-1928614, the National Eye Institute of the National Institutes of Health under Award Number R21EY033689, and DoD grant VR200130 under the “Delivering Sensory and Semantic Visual Information via Auditory Feedback on Mobile Technology”. C.F. is partially supported by NSF FW-HTF program under DUE-2026479. The content is solely the responsibility of the authors and does not necessarily represent the official views of the National Institutes of Health and NSF, and DoD.}


\informedconsent{Informed consent was obtained from all subjects involved in the study.}

\dataavailability{The data presented in this study are available in \href{https://drive.google.com/drive/folders/1xhGmdxgGzY0HCikQWW7MyAA2vF-MWyux?usp=sharing}{our google drive}} 

\conflictsofinterest{The authors declare no conflict of interest.} 

\begin{adjustwidth}{-\extralength}{0cm}

\reftitle{References}


\bibliography{UNav}

\begin{thebibliography}{999}

\bibitem[Kruk and Pate(2020)]{kruk2020lancet}
Kruk, M.E.; Pate, M.
\newblock The Lancet global health Commission on high quality health systems 1
  year on: progress on a global imperative.
\newblock {\em The Lancet global health} {\bf 2020}, {\em 8},~e30--e32.

\bibitem[Hakobyan \em{et~al.}(2013)Hakobyan, Lumsden, O’Sullivan, and
  Bartlett]{hakobyan2013mobile}
Hakobyan, L.; Lumsden, J.; O’Sullivan, D.; Bartlett, H.
\newblock Mobile assistive technologies for the visually impaired.
\newblock {\em Survey of ophthalmology} {\bf 2013}, {\em 58},~513--528.

\bibitem[Kandalan and Namuduri(2019)]{kandalan2019comprehensive}
Kandalan, R.N.; Namuduri, K.
\newblock A comprehensive survey of navigation systems for the visual impaired.
\newblock {\em arXiv preprint arXiv:1906.05917} {\bf 2019}.

\bibitem[Dakopoulos and Bourbakis(2009)]{dakopoulos2009wearable}
Dakopoulos, D.; Bourbakis, N.G.
\newblock Wearable obstacle avoidance electronic travel aids for blind: a
  survey.
\newblock {\em IEEE Transactions on Systems, Man, and Cybernetics, Part C
  (Applications and Reviews)} {\bf 2009}, {\em 40},~25--35.

\bibitem[Arandjelovic \em{et~al.}(2016)Arandjelovic, Gronat, Torii, Pajdla, and
  Sivic]{arandjelovic2016netvlad}
Arandjelovic, R.; Gronat, P.; Torii, A.; Pajdla, T.; Sivic, J.
\newblock NetVLAD: CNN architecture for weakly supervised place recognition.
\newblock In Proceedings of the Proceedings of the IEEE conference on computer
  vision and pattern recognition,  2016, pp. 5297--5307.

\bibitem[DeTone \em{et~al.}(2018)DeTone, Malisiewicz, and
  Rabinovich]{detone2018superpoint}
DeTone, D.; Malisiewicz, T.; Rabinovich, A.
\newblock Superpoint: Self-supervised interest point detection and description.
\newblock In Proceedings of the Proceedings of the IEEE conference on computer
  vision and pattern recognition workshops,  2018, pp. 224--236.

\bibitem[Manjari \em{et~al.}(2020)Manjari, Verma, and
  Singal]{manjari2020survey}
Manjari, K.; Verma, M.; Singal, G.
\newblock A survey on assistive technology for visually impaired.
\newblock {\em Internet of Things} {\bf 2020}, {\em 11},~100188.

\bibitem[Yang \em{et~al.}(2021)Yang, Yang, Wang, Zhou, Tian, and
  Li]{yang2021decimeter}
Yang, R.; Yang, X.; Wang, J.; Zhou, M.; Tian, Z.; Li, L.
\newblock Decimeter Level Indoor Localization Using WiFi Channel State
  Information.
\newblock {\em IEEE Sensors Journal} {\bf 2021}.

\bibitem[Al-Madani \em{et~al.}(2019)Al-Madani, Orujov, Maskeli{\=u}nas,
  Dama{\v{s}}evi{\v{c}}ius, and Ven{\v{c}}kauskas]{al2019fuzzy}
Al-Madani, B.; Orujov, F.; Maskeli{\=u}nas, R.; Dama{\v{s}}evi{\v{c}}ius, R.;
  Ven{\v{c}}kauskas, A.
\newblock Fuzzy logic type-2 based wireless indoor localization system for
  navigation of visually impaired people in buildings.
\newblock {\em Sensors} {\bf 2019}, {\em 19},~2114.

\bibitem[Feng and Kamat(2012)]{feng2012augmented}
Feng, C.; Kamat, V.R.
\newblock Augmented reality markers as spatial indices for indoor mobile AECFM
  applications.
\newblock In Proceedings of the Proceedings of 12th international conference on
  construction applications of virtual reality (CONVR 2012),  2012, pp.
  235--24.

\bibitem[Csurka \em{et~al.}(2004)Csurka, Dance, Fan, Willamowski, and
  Bray]{csurka2004visual}
Csurka, G.; Dance, C.; Fan, L.; Willamowski, J.; Bray, C.
\newblock Visual categorization with bags of keypoints.
\newblock In Proceedings of the Workshop on statistical learning in computer
  vision, ECCV. Prague,  2004, Vol.~1, pp. 1--2.

\bibitem[Sivic and Zisserman(2003)]{sivic2003video}
Sivic, J.; Zisserman, A.
\newblock Video Google: A text retrieval approach to object matching in videos.
\newblock In Proceedings of the Computer Vision, IEEE International Conference
  on. IEEE Computer Society,  2003, Vol.~3, pp. 1470--1470.

\bibitem[J{\'e}gou \em{et~al.}(2010)J{\'e}gou, Douze, Schmid, and
  P{\'e}rez]{jegou2010aggregating}
J{\'e}gou, H.; Douze, M.; Schmid, C.; P{\'e}rez, P.
\newblock Aggregating local descriptors into a compact image representation.
\newblock In Proceedings of the 2010 IEEE computer society conference on
  computer vision and pattern recognition. IEEE,  2010, pp. 3304--3311.

\bibitem[Torii \em{et~al.}(2015)Torii, Arandjelovic, Sivic, Okutomi, and
  Pajdla]{torii201524}
Torii, A.; Arandjelovic, R.; Sivic, J.; Okutomi, M.; Pajdla, T.
\newblock 24/7 place recognition by view synthesis.
\newblock In Proceedings of the Proceedings of the IEEE conference on computer
  vision and pattern recognition,  2015, pp. 1808--1817.

\bibitem[Bentley(1975)]{bentley1975multidimensional}
Bentley, J.L.
\newblock Multidimensional binary search trees used for associative searching.
\newblock {\em Communications of the ACM} {\bf 1975}, {\em 18},~509--517.

\bibitem[Gennaro \em{et~al.}(2001)Gennaro, Savino, and
  Zezula]{gennaro2001similarity}
Gennaro, C.; Savino, P.; Zezula, P.
\newblock Similarity search in metric databases through hashing.
\newblock In Proceedings of the Proceedings of the 2001 ACM workshops on
  Multimedia: multimedia information retrieval,  2001, pp. 1--5.

\bibitem[Philbin \em{et~al.}(2007)Philbin, Chum, Isard, Sivic, and
  Zisserman]{philbin2007object}
Philbin, J.; Chum, O.; Isard, M.; Sivic, J.; Zisserman, A.
\newblock Object retrieval with large vocabularies and fast spatial matching.
\newblock In Proceedings of the 2007 IEEE conference on computer vision and
  pattern recognition. IEEE,  2007, pp. 1--8.

\bibitem[Nister and Stewenius(2006)]{nister2006scalable}
Nister, D.; Stewenius, H.
\newblock Scalable recognition with a vocabulary tree.
\newblock In Proceedings of the 2006 IEEE Computer Society Conference on
  Computer Vision and Pattern Recognition (CVPR'06). Ieee,  2006, Vol.~2, pp.
  2161--2168.

\bibitem[Kendall \em{et~al.}(2015)Kendall, Grimes, and
  Cipolla]{kendall2015posenet}
Kendall, A.; Grimes, M.; Cipolla, R.
\newblock Posenet: A convolutional network for real-time 6-dof camera
  relocalization.
\newblock In Proceedings of the Proceedings of the IEEE international
  conference on computer vision,  2015, pp. 2938--2946.

\bibitem[Kendall and Cipolla(2017)]{kendall2017geometric}
Kendall, A.; Cipolla, R.
\newblock Geometric loss functions for camera pose regression with deep
  learning.
\newblock In Proceedings of the Proceedings of the IEEE conference on computer
  vision and pattern recognition,  2017, pp. 5974--5983.

\bibitem[Brahmbhatt \em{et~al.}(2018)Brahmbhatt, Gu, Kim, Hays, and
  Kautz]{brahmbhatt2018geometry}
Brahmbhatt, S.; Gu, J.; Kim, K.; Hays, J.; Kautz, J.
\newblock Geometry-aware learning of maps for camera localization.
\newblock In Proceedings of the Proceedings of the IEEE Conference on Computer
  Vision and Pattern Recognition,  2018, pp. 2616--2625.

\bibitem[Wang \em{et~al.}(2021)Wang, Xu, Ding, Huang, and Feng]{wang2021deep}
Wang, R.; Xu, X.; Ding, L.; Huang, Y.; Feng, C.
\newblock Deep Weakly Supervised Positioning for Indoor Mobile Robots.
\newblock {\em IEEE Robotics and Automation Letters} {\bf 2021}, {\em
  7},~1206--1213.

\bibitem[Kendall and Cipolla(2017)]{Kendall_2017_CVPR}
Kendall, A.; Cipolla, R.
\newblock Geometric Loss Functions for Camera Pose Regression With Deep
  Learning.
\newblock In Proceedings of the Proceedings of the IEEE Conference on Computer
  Vision and Pattern Recognition (CVPR),  2017.

\bibitem[Arth \em{et~al.}(2015)Arth, Pirchheim, Ventura, Schmalstieg, and
  Lepetit]{arth2015instant}
Arth, C.; Pirchheim, C.; Ventura, J.; Schmalstieg, D.; Lepetit, V.
\newblock Instant outdoor localization and slam initialization from 2.5 d maps.
\newblock {\em IEEE Transactions on Visualization \& Computer Graphics} {\bf
  2015}, {\em 21},~1309--1318.

\bibitem[Song \em{et~al.}(2016)Song, Chen, Wang, Zhang, and Li]{song20166}
Song, Y.; Chen, X.; Wang, X.; Zhang, Y.; Li, J.
\newblock 6-DOF image localization from massive geo-tagged reference images.
\newblock {\em IEEE Transactions on Multimedia} {\bf 2016}, {\em
  18},~1542--1554.

\bibitem[Liu \em{et~al.}(2017)Liu, Li, and Dai]{liu2017efficient}
Liu, L.; Li, H.; Dai, Y.
\newblock Efficient global 2d-3d matching for camera localization in a
  large-scale 3d map.
\newblock In Proceedings of the Proceedings of the IEEE International
  Conference on Computer Vision,  2017, pp. 2372--2381.

\bibitem[Sv{\"a}rm \em{et~al.}(2016)Sv{\"a}rm, Enqvist, Kahl, and
  Oskarsson]{svarm2016city}
Sv{\"a}rm, L.; Enqvist, O.; Kahl, F.; Oskarsson, M.
\newblock City-scale localization for cameras with known vertical direction.
\newblock {\em IEEE transactions on pattern analysis and machine intelligence}
  {\bf 2016}, {\em 39},~1455--1461.

\bibitem[Taira \em{et~al.}(2018)Taira, Okutomi, Sattler, Cimpoi, Pollefeys,
  Sivic, Pajdla, and Torii]{taira2018inloc}
Taira, H.; Okutomi, M.; Sattler, T.; Cimpoi, M.; Pollefeys, M.; Sivic, J.;
  Pajdla, T.; Torii, A.
\newblock InLoc: Indoor visual localization with dense matching and view
  synthesis.
\newblock In Proceedings of the Proceedings of the IEEE Conference on Computer
  Vision and Pattern Recognition,  2018, pp. 7199--7209.

\bibitem[Toft \em{et~al.}(2018)Toft, Stenborg, Hammarstrand, Brynte, Pollefeys,
  Sattler, and Kahl]{toft2018semantic}
Toft, C.; Stenborg, E.; Hammarstrand, L.; Brynte, L.; Pollefeys, M.; Sattler,
  T.; Kahl, F.
\newblock Semantic match consistency for long-term visual localization.
\newblock In Proceedings of the Proceedings of the European Conference on
  Computer Vision (ECCV),  2018, pp. 383--399.

\bibitem[Sarlin \em{et~al.}(2019)Sarlin, Cadena, Siegwart, and
  Dymczyk]{sarlin2019coarse}
Sarlin, P.E.; Cadena, C.; Siegwart, R.; Dymczyk, M.
\newblock From coarse to fine: Robust hierarchical localization at large scale.
\newblock In Proceedings of the Proceedings of the IEEE/CVF Conference on
  Computer Vision and Pattern Recognition,  2019, pp. 12716--12725.

\bibitem[Sumikura \em{et~al.}(2022)Sumikura, Shibuya, and
  Sakurada]{sumikura2022openvslam}
Sumikura, S.; Shibuya, M.; Sakurada, K.
\newblock OpenVSLAM: a versatile visual SLAM framework.
\newblock {\em ACM SIGMultimedia Records} {\bf 2022}, {\em 11},~1--1.

\bibitem[Mur-Artal and Tard{\'o}s(2017)]{mur2017orb}
Mur-Artal, R.; Tard{\'o}s, J.D.
\newblock Orb-slam2: An open-source slam system for monocular, stereo, and
  rgb-d cameras.
\newblock {\em IEEE transactions on robotics} {\bf 2017}, {\em 33},~1255--1262.

\bibitem[Rublee \em{et~al.}(2011)Rublee, Rabaud, Konolige, and
  Bradski]{rublee2011orb}
Rublee, E.; Rabaud, V.; Konolige, K.; Bradski, G.
\newblock ORB: An efficient alternative to SIFT or SURF.
\newblock In Proceedings of the 2011 International conference on computer
  vision. Ieee,  2011, pp. 2564--2571.

\bibitem[Sch{\"o}nberger \em{et~al.}(2016)Sch{\"o}nberger, Zheng, Frahm, and
  Pollefeys]{schonberger2016pixelwise}
Sch{\"o}nberger, J.L.; Zheng, E.; Frahm, J.M.; Pollefeys, M.
\newblock Pixelwise view selection for unstructured multi-view stereo.
\newblock In Proceedings of the European Conference on Computer Vision.
  Springer,  2016, pp. 501--518.

\bibitem[Schonberger and Frahm(2016)]{schonberger2016structure}
Schonberger, J.L.; Frahm, J.M.
\newblock Structure-from-motion revisited.
\newblock In Proceedings of the Proceedings of the IEEE conference on computer
  vision and pattern recognition,  2016, pp. 4104--4113.

\bibitem[Sarlin \em{et~al.}(2020)Sarlin, DeTone, Malisiewicz, and
  Rabinovich]{sarlin2020superglue}
Sarlin, P.E.; DeTone, D.; Malisiewicz, T.; Rabinovich, A.
\newblock Superglue: Learning feature matching with graph neural networks.
\newblock In Proceedings of the Proceedings of the IEEE/CVF conference on
  computer vision and pattern recognition,  2020, pp. 4938--4947.

\bibitem[Rizzo(2017)]{rizzo2017somatosensory}
Rizzo, J.R.
\newblock Somatosensory feedback wearable object,  2017.
\newblock US Patent 9,646,514.

\bibitem[Niu \em{et~al.}(2017)Niu, Qian, Rizzo, Hudson, Li, Enright, Sperling,
  Conti, Wong, and Fang]{niu2017wearable}
Niu, L.; Qian, C.; Rizzo, J.R.; Hudson, T.; Li, Z.; Enright, S.; Sperling, E.;
  Conti, K.; Wong, E.; Fang, Y.
\newblock A wearable assistive technology for the visually impaired with door
  knob detection and real-time feedback for hand-to-handle manipulation.
\newblock In Proceedings of the Proceedings of the IEEE International
  Conference on Computer Vision Workshops,  2017, pp. 1500--1508.

\bibitem[Shoureshi \em{et~al.}(2017)Shoureshi, Rizzo, and
  Hudson]{shoureshi2017smart}
Shoureshi, R.A.; Rizzo, J.R.; Hudson, T.E.
\newblock Smart wearable systems for enhanced monitoring and mobility.
\newblock In Proceedings of the Advances in Science and Technology. Trans Tech
  Publ,  2017, Vol. 100, pp. 172--178.

\bibitem[Denis and Daniele(2004)]{denis2004nlos}
Denis, B.; Daniele, N.
\newblock NLOS ranging error mitigation in a distributed positioning algorithm
  for indoor UWB ad-hoc networks.
\newblock In Proceedings of the International Workshop on Wireless Ad-Hoc
  Networks, 2004. IEEE,  2004, pp. 356--360.

\bibitem[Hsu \em{et~al.}(2015)Hsu, Chen, and Kamijo]{hsu2015evaluation}
Hsu, L.T.; Chen, F.; Kamijo, S.
\newblock Evaluation of multi-GNSSs and GPS with 3D map methods for pedestrian
  positioning in an urban canyon environment.
\newblock {\em IEICE Transactions on Fundamentals of Electronics,
  Communications and Computer Sciences} {\bf 2015}, {\em 98},~284--293.

\bibitem[Pan \em{et~al.}(2022)Pan, Pollefeys, and Larsson]{pan2022camera}
Pan, L.; Pollefeys, M.; Larsson, V.
\newblock Camera Pose Estimation Using Implicit Distortion Models.
\newblock In Proceedings of the Proceedings of the IEEE/CVF Conference on
  Computer Vision and Pattern Recognition,  2022, pp. 12819--12828.

\bibitem[Chen \em{et~al.}(2022)Chen, Qian, and Furukawa]{chen2022heat}
Chen, J.; Qian, Y.; Furukawa, Y.
\newblock HEAT: Holistic Edge Attention Transformer for Structured
  Reconstruction.
\newblock In Proceedings of the Proceedings of the IEEE/CVF Conference on
  Computer Vision and Pattern Recognition,  2022, pp. 3866--3875.

\end{thebibliography}

\end{adjustwidth}
\end{document}